\def\year{2019}\relax
\documentclass[letterpaper]{article} %DO NOT CHANGE THIS
\usepackage{aaai19}  %Required
\usepackage{amsmath,amssymb}
\usepackage{times}  %Required
\usepackage{helvet}  %Required
\usepackage{multirow}
\usepackage{courier}  %Required
\newcommand{\eg}{e.g.}
\newcommand{\ie}{i.e.}
\newcommand{\etal}{et al.}
\newcommand{\tuple}[3]{$\langle$\emph{#1-#2-#3}$\rangle$}
\usepackage{url}  %Required
\usepackage{graphicx}  %Required
\begin{document}
% The file aaai.sty is the style file for AAAI Press 
% proceedings, working notes, and technical reports.
%
\title{Learning to Steer by Mimicking Features from Heterogeneous Auxiliary Networks}
\author{Yuenan Hou$^{1}$, Zheng Ma$^{2}$, Chunxiao Liu$^{2}$, and Chen Change Loy$^{3}$\\
$^{1}$The Chinese University of Hong Kong $^{2}$SenseTime Group Limited $^{3}$Nanyang Technological University\\
hy117@ie.cuhk.edu.hk, \{mazheng, liuchunxiao\}@sensetime.com, ccloy@ntu.edu.sg
}
\maketitle
\def\algorithmname{FM-Net}

\begin{abstract}

The training of many existing end-to-end steering angle prediction models heavily relies on steering angles as the supervisory signal. Without learning from much richer contexts, these methods are susceptible to the presence of sharp road curves, challenging traffic conditions, strong shadows, and severe lighting changes. 
In this paper, we considerably improve the accuracy and robustness of predictions through \textit{heterogeneous auxiliary networks feature mimicking}, a new and effective training method that provides us with much richer contextual signals apart from steering direction. Specifically, we train our steering angle predictive model by distilling multi-layer knowledge from multiple heterogeneous auxiliary networks that perform related but different tasks, e.g., image segmentation or optical flow estimation. As opposed to multi-task learning, our method does not require expensive annotations of related tasks on the target set. This is made possible by applying contemporary off-the-shelf networks on the target set and mimicking their features in different layers after transformation. The auxiliary networks are discarded after training without affecting the runtime efficiency of our model.
Our approach achieves a new state-of-the-art on Udacity and Comma.ai, outperforming the previous best by a large margin of \textbf{12.8}\% and \textbf{52.1}\%\footnote{We compare the mean absolute error against the top entry `komanda' on Udacity leaderboard~\cite{udacity}, which uses a 3D CNN with LSTM.}, respectively. Encouraging results are also shown on Berkeley Deep Drive (BDD) dataset.

\end{abstract}

%////////////////////////////////
\section{Introduction}
\label{sec:introduction}
%////////////////////////////////
% !TEX root = ../deepdrive.tex

Autonomous driving is conventionally formulated and solved as a collection of sub-problems, including perception, decision, path planning, and control~\cite{paden2016survey}. Recent approaches address the problem in an end-to-end manner, in which a convolutional neural network (CNN) is trained end-to-end to map raw visual observations (\eg, images or videos) obtained from a single front-facing camera directly to steering commands~\cite{bojarski2016end}.

\begin{figure}[t]
\begin{center}
\includegraphics[width=\linewidth]{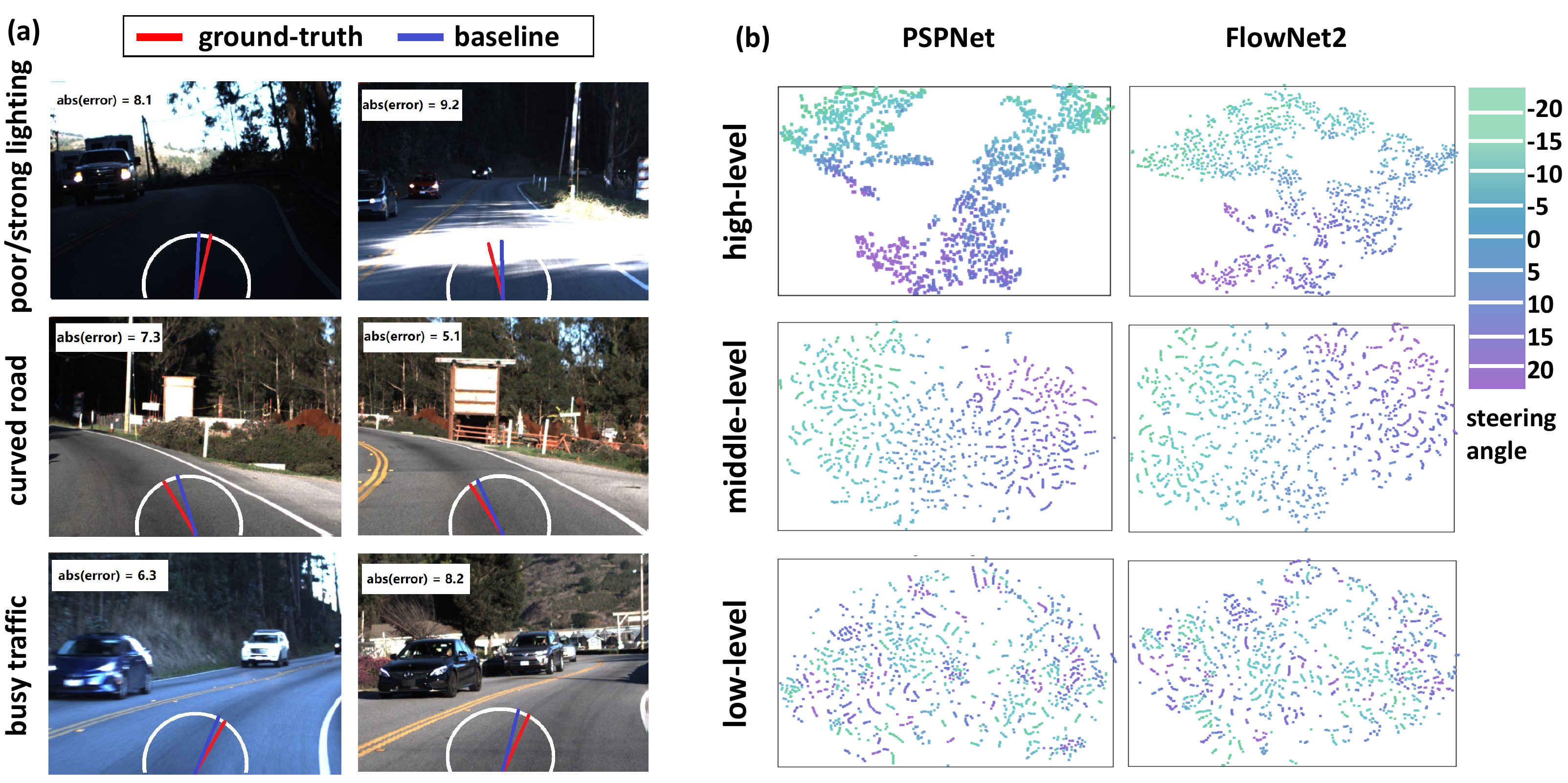}
\caption{\textbf{(a) Failure cases of baseline model.} A baseline model (3D CNN + LSTM)~\cite{udacity} that is trained with steering angle alone tends to fail in challenging cases. \textbf{(b) Deep feature embeddings of auxiliary networks.} The embedding of features extracted from PSPNet~\cite{zhao2017pyramid} and FlowNet2~\cite{ilg2017flownet}.  Each point corresponds to a deep feature vector of a video frame. Every point is encoded by colour so that points to the most positive steering angle are lilac and points with the most negative steering angle are aqua. As can be observed, features at different layers are highly indicative of steering angle prediction.}
\label{fig:overview}
\end{center}
\vspace{-0.7cm}
\end{figure}

Steering angle is often used as the sole supervisory signal for training a network~\cite{bojarski2016end,pomerleau1989alvinn}. Some studies improve the training by multi-task learning~\cite{chowdhuri2017multi,yang2018end}, \ie, requiring the network to predict additional labels such as vehicle speed and steering torque.
These supervisory signals are informative but do not ensure effective representation learning to capture rich environmental contexts, \eg, physical scene constraints or coexistence of scene objects,  which are crucial for driving.
Without such spatial and object awareness, existing methods often fail in challenging cases that involve severe lighting changes, strong shadows, sharp turns, or busy traffic. In Fig.~\ref{fig:overview}(a), we show some of the challenging cases on which a baseline fails.

A plausible way to solve the problem above is by widening the scope of multi-task learning from speed or torque to more complex tasks such as scene segmentation, lane detection, or optical flow estimation. These tasks capture scene structure and object motion that likely benefit steering angle prediction. However, introducing side tasks for multi-task learning requires one to collect extra task-specific annotations for the target scene, a process that is both laborious and expensive.
An alternative approach is to pre-train a network with related tasks such as scene segmentation and fine-tune the model to the steering angle prediction task. This method relaxes the need of target scene annotations since pre-training can exploit data collected from a different scene. Our experiments, however, show that this indirect approach only improves steering angle prediction marginally.

In this study, we train our model, \textbf{\algorithmname}, with a new and effective technique that brings drastic improvement to the performance of end-to-end steering angle prediction. %, by as much as \tofill{x.x}\% compared to the previous best~\misscite on Udacity and \tofill{x.x}\% on Comma.ai as compared to~\misscite.
Unlike multi-task learning, our method does not require additional annotations of side tasks on target scene.
This is made possible by drawing inspiration from Hinton \etal's seminal work on knowledge distillation~\cite{hinton2015distilling}.
In contrast to \cite{hinton2015distilling} that distils knowledge in an ensemble of large models into a single small model, we propose `\textit{heterogeneous auxiliary networks feature mimicking}', which allows the learning of a steering angle predictive model by distilling knowledge from heterogeneous off-the-shelf networks.
Specifically, there are many strong and state-of-the-art networks such as PSPNet~\cite{zhao2017pyramid} for segmentation, and FlowNet2~\cite{ilg2017flownet} for optical flow estimation.
Applying these networks on the target data can generate features that are highly indicative of the final prediction of steering angles. This is evident from the embedding of these features shown in Fig.~\ref{fig:overview}(b). As can be observed, both high- and low-layer features are structured and meaningful.
Clearly, the high and middle levels contain direct hints for angle prediction. Low-level features are more scattered but they still show well-clustered embeddings. %The usefulness of features from different levels will be examined empirically in our experiments.

In this work, we aim to train a more accurate steering angle predictive model by enforcing it to approximate the multi-layer representation of those well-established auxiliary networks.
In contrast to~\cite{hinton2015distilling}, we find an entirely different application of mimicking heterogeneous networks for learning rich contexts.
In addition, we explore the use of deep features extracted from different layers of auxiliary networks as targets beyond the logits (the inputs to the final softmax) as proposed by~\cite{hinton2015distilling}.
We summarize the \textbf{contributions} of this paper as follows:

\vspace{0.1cm}
\noindent
1) We present an effective training method that drastically improves the performance of end-to-end steering angle prediction through mimicking features from well-established and cheap-to-access auxiliary networks. The auxiliary network is only used in the training stage and brings no computation cost during the deployment.

\noindent
2) We demonstrate through extensive and systematic experiments that mimicking can be conducted simultaneously from heterogeneous auxiliary networks. In addition, mimicking can be performed at multiple layers of an auxiliary network while still benefiting the main network. This allows our main network to acquire rich contexts of different natures and spatial resolutions.

\noindent
3) The mimicking process is non-trivial as the original features extracted from different layers of auxiliary networks are high-dimensional.
We show effective ways of pooling these features to a lower dimension for regularizing and training a 3D CNN for steering angle prediction.

Apart from auxiliary network mimicking, we show that both network choice and initialization play a crucial role in prediction performance. The deepest network in the literature is fewer than 10 layers~\cite{bojarski2016end}. We advance the state-of-the-art by proposing a 50-layer 3D ResNet. We inflate our 3D convolutional network from a 2D ImageNet model. This initialization scheme, which was originally proposed for action recognition~\cite{carreira2017quo}, provides us with a strong ResNet-based model for representation learning and mimicking.
Extensive experiments are conducted on two public datasets, namely, Udacity~\cite{udacity} and Comma.ai~\cite{santana2016learning}. Our method surpasses previous methods by a large margin and records a new state-of-the-art in steering angle prediction, with a mean absolute error (MAE) of \textbf{1.62} and \textbf{0.70} and root mean square error (RMSE) of \textbf{2.35} and \textbf{0.98}. 

%////////////////////////////////
\section{Related Work}
\label{sec:related_work}
%////////////////////////////////
% !TEX root = ../deepdrive.tex

\noindent
\textbf{End-to-end learning for self driving}.
Learn-to-steer with end-to-end optimization was first demonstrated in~\cite{pomerleau1989alvinn}. The study utilized a shallow neural network to predict actuation from images. Although the driving conditions were quite simple, its appealing performance showcased the possibility of applying neural networks to autonomous driving. A similar idea is later presented by exploiting a deep CNN to output steering commands from images~\cite{bojarski2016end}. Due to its good prediction results achieved in highway driving, the structure (a network with a normalization layer, five convolutional layers, and three fully connected layers) has become the base model for many studies~\cite{xu2017end,chi2017deep}.
We found that there is a lack of new models publicly available for our problem. Most of the models are fewer than 10 layers and do not adopt a contemporary architecture like ResNet~\cite{he2016deep}. In this study, we contribute a 50-layer 3D ResNet model for steering angle prediction and make it available to the research community\footnote{Code is available at \emph{https://cardwing.github.io/projects/FM-Net}}.

\begin{figure}[t]
  \centering
  \includegraphics[width=\linewidth]{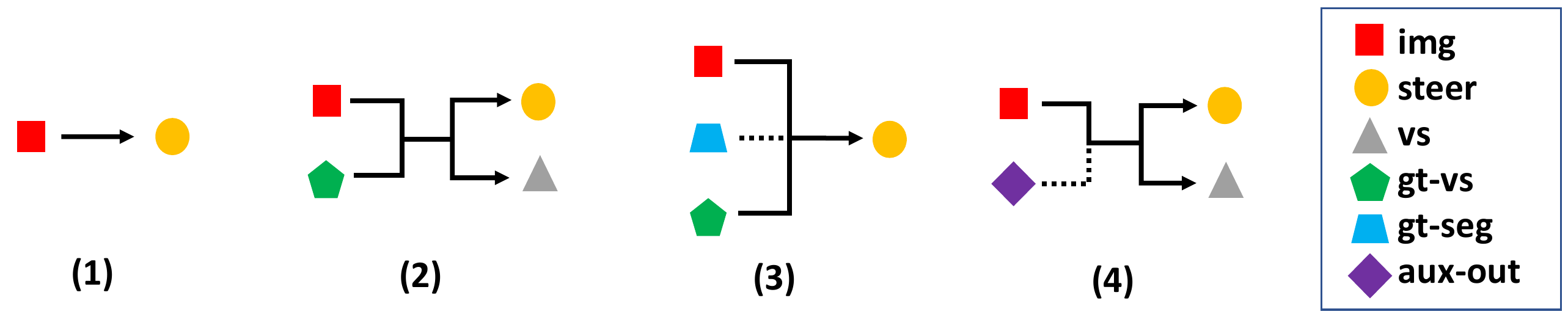}
  \vskip -0.3cm
  \caption{\textbf{Four main paradigms of learning steering angle prediction}. Each subfigure shows the training sources and prediction tasks. Six abbreviations in the legend denote image sequence, steering angle, vehicle state, ground-truth vehicle state, ground-truth segmentation labels and output features of auxiliary networks, respectively. The last paradigm 4 is proposed in this paper. A dotted line means that the corresponding source is dropped after training.}
\label{fig:paradigms}
\end{figure}

Currently, there are four main paradigms adopted for training an end-to-end steering angle prediction model (see Fig.~\ref{fig:paradigms}). A direct mapping of image sequences to steering angles is the simplest form.
Researchers soon found that using steering commands as the sole supervisory signal is too weak to train deep networks. Some studies incorporated multi-task learning in the pipeline~\cite{chi2017deep,xu2017end,yang2018end,chowdhuri2017multi} to prevent over-fitting and improved the prediction accuracy of steering angles. The second and third paradigms in Fig.~\ref{fig:paradigms} correspond to these variants.
For instance, Chowdhuri \etal~\cite{chowdhuri2017multi} proposed a light-weight Z2Color network and executed steering commands by utilizing vehicle-state indicators (behavioral modalities) as secondary input data besides image sequences.
To supplement visual inputs, in~\cite{yang2018end} the steering prediction and speed prediction were simultaneously learnt in the framework of multi-task learning.
To leverage auxiliary tasks for feature learning, Xu \etal~\cite{xu2017end} developed a FCN-LSTM framework, which learnt jointly from steering loss and image segmentation loss. However, this method needs extra labels for the auxiliary segmentation task. Our proposed framework learns directly from off-the-shelf models without any extra labeling for target scene apart from the ground-truth steering angles. It belongs to the last paradigm in Fig.~\ref{fig:paradigms}. The main difference between our scheme and multi-task learning is that extra labels are not needed during training and image sequences are the sole input of the network.

\vspace{0.1cm}
\noindent
\textbf{Network and feature mimicking}.
Network mimicking is originally introduced for small networks to distil knowledge from an ensemble of large networks for network acceleration and compression~\cite{hinton2015distilling,ba2014deep} by forcing a small network to mimic outputs of large networks. In~\cite{romero2014fitnets} the mimicking idea was applied in image classification, where a student network was required to learn the intermediate output of a teacher network. This mimicking strategy is also known as feature mimicking. Li \etal~\cite{li2017mimicking} further extended feature mimicking into object detection tasks, where a small network was used to mimic spatially sampled features of large networks. Despite the progress made in feature mimicking, existing mimicking approaches are limited to transferring knowledge from large networks to a small network, and these networks share a common task. In \cite{Gupta2016distillation}, feature mimicking was used to teach a new CNN for a new image modality (like depth images), by teaching the network to reproduce the mid-level semantic representations learned from a well labeled image modality.
In this work, we demonstrate the possibility of transferring knowledge between large networks that perform \textit{heterogeneous} tasks. We also explore feature mimicking at different layers of a teacher network to distil richer information for learning.

%////////////////////////////////
\section{Methodology}
\label{sec:methodology}
%////////////////////////////////
% !TEX root = ../deepdrive.tex

%We first describe a general formulation of steering angle prediction. We then present the proposed approach for learning from heterogeneous auxiliary networks. Finally, we discuss methods that can be used to improve network initialization and training.

%-----------------------------------
\subsection{Preliminary}
\label{subsec:problem_formulation}
%-----------------------------------

The general objective of steering angle prediction is to predict the angle $p$ given a video frame $x$.
Typically, one would use a video clip, $\textbf{x} = (x_{1}, x_{2}, \dots, x_{N})$ as input to encapsulate the temporal information, and then learn a function $\mathcal{F}: \textbf{x} \mapsto p$ for prediction.
%
%\begin{equation}
%(p_{1}, p_{2}, ..., p_{N}) = F(x_{1}, x_{2}, ..., x_{N}; \Theta),
%\end{equation}
%
Recent studies use convolutional networks as $\mathcal{F}$ for end-to-end prediction.
Multi-task learning~\cite{chowdhuri2017multi,yang2018end} assumes additional targets apart from steering angles. Examples of such targets include speed of the vehicle, steering wheel torque, or GPS trajectory. It can also take a richer form such as a sequence of scene segmentation maps. Here, we use $b_l$ to denote the $l$-th additional target label.
The function becomes $\mathcal{F}: \textbf{x} \mapsto (p, \{b_l\}^L_{l=1})$, where $L$ is the number of additional tasks.

In this study, we follow existing practices to train the proposed \algorithmname~to make prediction on steering angle, speed of the vehicle, and steering wheel torque. The ground-truth and metadata are readily available from many benchmark datasets such as Udacity~\cite{udacity} and Comma.ai~\cite{santana2016learning}. The key difference of our work is that we regularize the learning of our model by requiring it to approximate the features extracted from different layers of heterogeneous auxiliary networks.
We introduce our method in the next section.

%-----------------------------------
\subsection{Multi-layer Feature Mimicking from Heterogeneous Networks}
\label{subsec:our_approach}
%-----------------------------------

\begin{figure*}[t]
  \centering
  \includegraphics[width=0.9\linewidth]{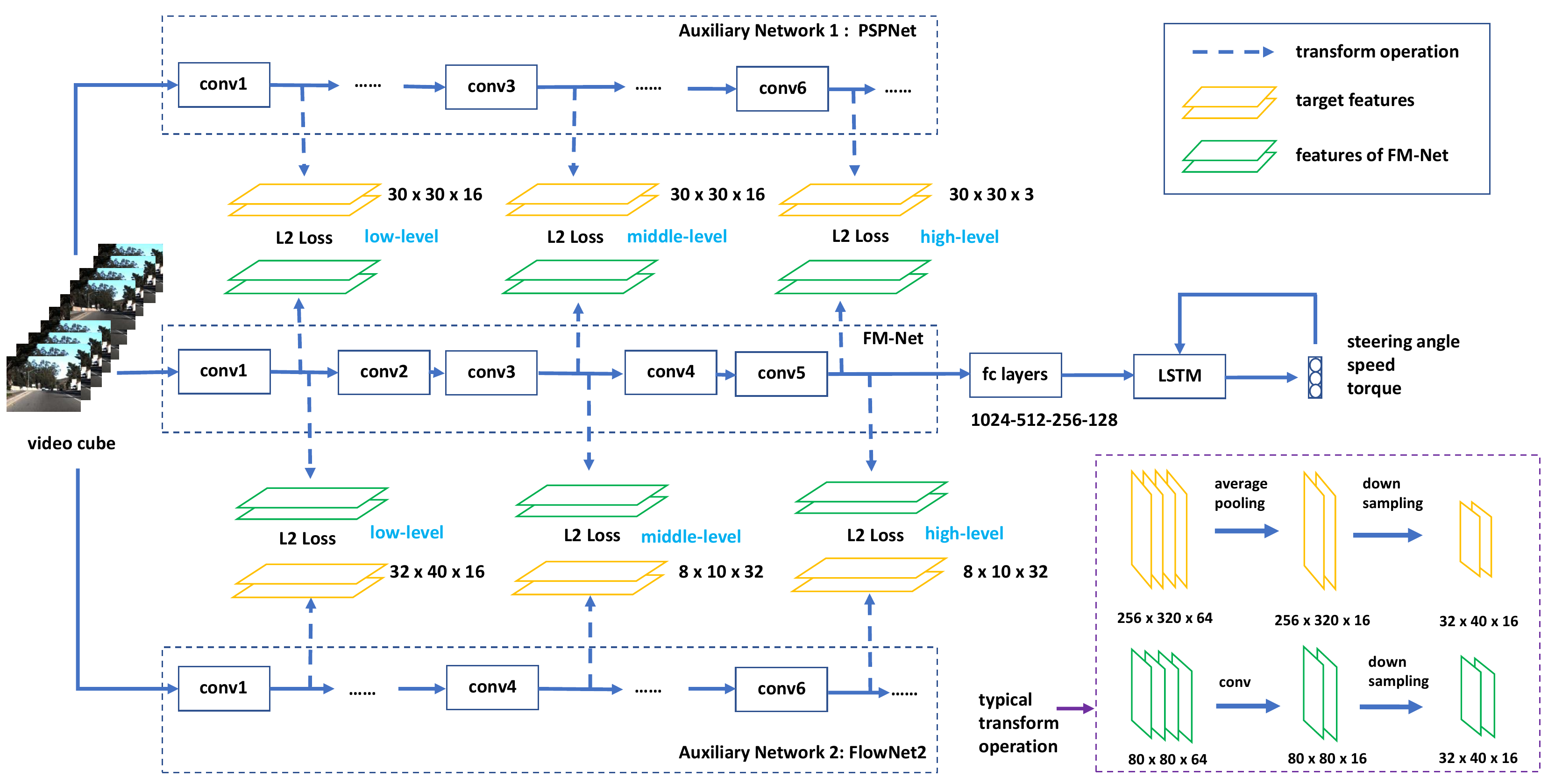}
  \vskip -0.2cm
  \caption{\textbf{Feature Mimicking from Heterogeneous Auxiliary Network.} The main network is \algorithmname~that takes an architecture of 3D ResNet + LSTM. The two auxiliary networks are PSPNet~\cite{zhao2017pyramid} and FlowNet2~\cite{ilg2017flownet,Hui_2018_CVPR}. Transformation layers are introduced to ensure a low yet compatible spatial dimension for feature mimicking at \tuple{low}{middle}{high} mimicking paths. The input is a sequence of images and the prediction outputs are steering angles, speed of vehicle, and steering wheel torque. The details of the architecture and mimicking paths are shown in Table~\ref{tab:feature_dimension}.}
  \label{fig:framework}
\end{figure*}

We first provide an overview of our framework that is shown in Fig.~\ref{fig:framework}.
We denote the main network \algorithmname~as $M$. It is a 50-layer 3D ResNet with a Long-Short-Term Memory (LSTM) module~\cite{sak2014long}.
We provide architectural details of this network in the next subsection.
At each time step, a sequence of $N$ video frames $\textbf{x}$ is fed to the main network. Fully-connected layers are introduced after the convolutional layers to transform the feature maps to a  compact feature vector. To further capture temporal dynamics, which is crucial for smooth angle prediction, a LSTM module is added thereafter.
Three vehicle state indicators, namely steering angle, speed of the vehicle, and steering wheel torque are predicted according to the extracted feature vectors and predicted vehicle state indicators of the last time step. These previous vehicle states will contribute to the prediction of our network.
We next detail our approach on heterogeneous feature mimicking.

\begin{figure}[t]
  \centering
  \includegraphics[width=\linewidth]{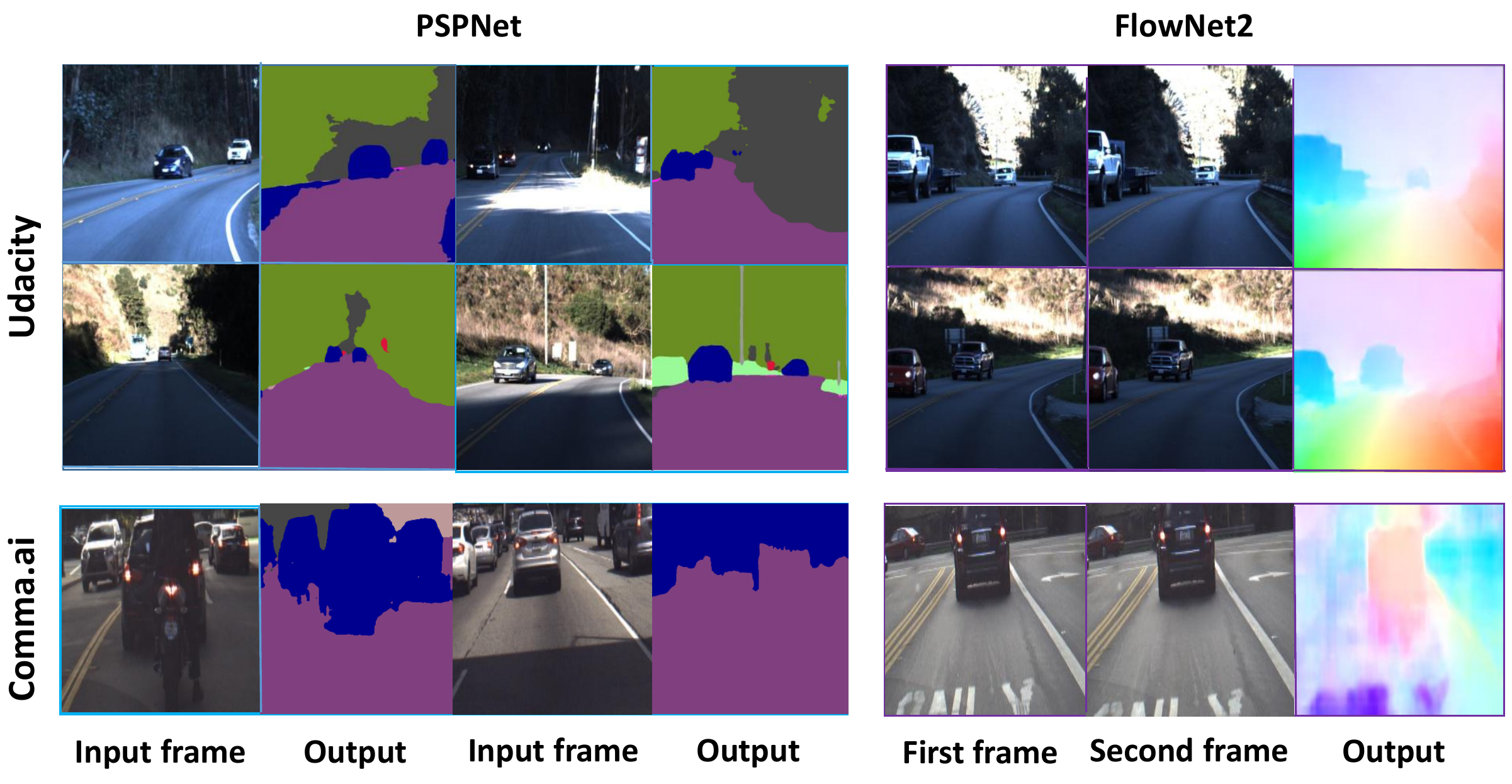}
  \vskip -0.2cm
  \caption{Auxiliary networks PSPNet and FlowNet2 show satisfactory generalization on unseen target data. }
  \label{fig:aux_gen}
  \vspace{-0.2cm}
\end{figure}

\vspace{0.1cm}
\noindent
\textbf{Auxiliary networks}.
In our approach, apart from the main network, we assume a set of $K$ heterogeneous auxiliary networks $\mathcal{A} = \{ A_1, A_2 \dots A_K \} $ during training stage. These auxiliary networks are chosen from off-the-shelf networks that achieve strong performance in their respective task, \eg, PSPNet~\cite{zhao2017pyramid} for image segmentation and FlowNet2~\cite{ilg2017flownet,Hui_2018_CVPR} optical flow estimation. In this study, we use two auxiliary networks, \ie, PSPNet and FlowNet2. They represent a well-suited choice of auxiliary networks as PSPNet captures the scene semantical structure while FlowNet2 encapsulates information on moving objects. It is interesting to see that both networks have strong generalization capability when they are applied on the unseen target data, as shown in Fig.~\ref{fig:aux_gen} and Fig.~\ref{fig:overview}(b).
Mimicking features from both of these networks contribute to performance improvement of the main network, as we will show in our experiments.
Note that although we show two auxiliary networks in our study, one can easily generalize to more networks. These auxiliary networks will be discarded after training.

\begin{table}[t]
\centering
\caption{The different dimensions ($w \times w \times c$, where $w \times w$ is the size of feature maps and $c$ is the number of channels) produced by the transformation layers at \tuple{low}{middle}{high} mimicking paths.
Here, the output dimension of PSPNet is the same for Udacity and Comma.ai, but different on FlowNet because PSPNet normalizes the input image to the same resolution so there is no discrepancy of output dimensions on Udacity and Comma.ai.}
\label{tab:feature_dimension}
\begin{tabular}{c|c|c|c}
\hline
Auxiliary & Mimicking & \multicolumn{2}{|c}{Output Dimension ($w \times w \times c$)} \\
\cline{3-4}
Networks & Path & Udacity & Comma.ai \\
\hline \hline
\multirow{3}*{FlowNet} & Low & 32 $\times$ 40 $\times$ 16 & 12 $\times$ 20 $\times$ 32 \\
\cline{2-4}
~ & Middle & 8 $\times$ 10 $\times$ 32 & 12 $\times$ 20 $\times$ 32 \\
\cline{2-4}
~ & High & 8 $\times$ 10 $\times$ 32 & 3 $\times$ 5 $\times$ 64 \\
\hline
\multirow{3}*{PSPNet} & Low & \multicolumn{2}{|c}{ 30 $\times$ 30 $\times$ 16} \\
\cline{2-4}
~ & Middle & \multicolumn{2}{|c}{ 30 $\times$ 30 $\times$ 16 } \\
\cline{2-4}
~ & High & \multicolumn{2}{|c}{ 30 $\times$ 30 $\times$ 3} \\
\hline
\end{tabular}
\end{table}

Given an auxiliary network $A_k$, we can extract features from its different feature layers. Formerly, we denote features extracted from $j$-th layer of $A_k$ as $\textbf{f}_{jk}$. We use $N_{A_k}$ to denote the number of layers from which features are extracted from $A_k$.
Here, $\textbf{f}_{jk}$ can be feature maps (output of convolution layers) or feature vectors (output of fully connected layers).
To perform multi-layer feature mimicking, we pair each of the $N_{A_k}$ layers of $A_k$ with a corresponding layer of the main network $M$. Similar to the way we obtain $\textbf{f}_{jk}$ from the auxiliary network, we  extract features at designated layers of the main network and obtain features $\textbf{e}_j$ for its $j$-layer. To examine the usefulness of different layers more conveniently, we assume three paired levels \tuple{low}{middle}{high} from both the main network and auxiliary networks.
These three levels are chosen according to the network depth (for high-level mimicking) and the size of receptive fields (for low- and middle-level mimicking). While our current design is shown effective, this is by no means the only option. More levels can be attempted. With different designs of transformation layers (described next), features of different scales are applicable too.
%
%These three levels roughly correspond to different parts of a network that transit from `general' to `specific'~\cite{yosinski2014transferable}. The more `specific' a layer is, its features are more tailored for its training task.
%
In the experimental section, we will systematically examine the benefits of mimicking different individual level of features.

\vspace{0.1cm}
\noindent
\textbf{Transformation layers}.
We need to address two issues during the process of feature mimicking.
Firstly, the source features generated by auxiliary networks are of high dimensionality. For instance, the output dimension of the first convolution block of PSPNet is $179 \times 179 \times 128$. Simply approximating the features would cause difficulty in training the main network.
Secondly, we wish to retain the spatial information encoded in the feature maps extracted from both main network and auxiliary networks. The spatial information is crucial to provide contextual information of a driving scene.

To address the two issues, we propose to insert two transformation layers between each of the \tuple{low}{middle}{high} mimicking paths, as shown in Fig.~\ref{fig:framework}.
A transformation layer $\Phi$ is designated for the main network, whilst another type of transformation layer $\Psi$ is designed for auxiliary networks.
The common goals of these transformation layers are (1) to reduce the dimensionality of the original feature maps, and (2) to retain sufficient spatial information of the original feature maps.
The parameters in these transformation layers are learned in the training process and add little extra computation to training.

For the transformation layer $\Phi$ of main network, we use $1 \times 1$ convolution operation to reduce the number of channels and an upsampling/downsampling operation to alter the size of features.
For the transformation layer $\Psi$ of auxiliary networks, we employ an average pooling operation to reduce the number of channels and an upsampling/downsampling operation.
If the sizes of main network's features or auxiliary networks' features are the same as the targeted sizes, then we do not perform upsampling/downsampling operation.
Note that average pooling is used instead of $1 \times 1$ convolution because we need a deterministic target for training the FM-Net. Average pooling meets this purpose while compacting the redundant features of auxiliary networks.

%
%The general rule of thumb is not to flatten the original feature maps into a vector so that spatial information can be kept.

%For high-level features of auxiliary networks which are to be mimicked, only semantic-related feature channels are selected (see Fig. \ref{fig6}). More specifically, the output digits (before softmax layer) of PSPNet are selected and resized to 30 $\times$ 30 to facilitate learning. Besides, since the number of predicted classes of PSPNet is 19, we further reduce the dimension of training samples by only choosing three classes, i.e., the road, vehicle and all stuff. Therefore, the high-level features PSPNet provides have the shape of 30 $\times$ 30 $\times$ 3 (see Table \ref{tab1}).

\vspace{0.1cm}
\noindent
\textbf{Loss}.
The overall loss comprises of three terms:
\begin{equation}
\label{eqn:final_loss}
\begin{split}
\mathcal{L} = & \underbrace{\mathcal{L}_\mathrm{steer}(p, \hat{p})}_{\text{steering loss}} + \underbrace{\sum\nolimits_{l=1}^{L} \alpha_{l} \mathcal{L}_\mathrm{multi}(b_{l}, \hat{b}_{l})}_{\text{multi-task loss}}
\\
& + \underbrace{\sum\nolimits_{k=1}^{K} \sum\nolimits_{j=1}^{N_{A_k}} \beta_{k} \mathcal{L}_\mathrm{mimic}(\Phi(\textbf{e}_{j}), \Psi(\textbf{f}_{jk}))}_{\text{mimicking loss}}).
\end{split}
\end{equation}

The first term is the steering angle prediction loss, which is typically defined as a L2 loss, and $\hat{p}$ are angle predictions produced by the main network.
The second term is the multi-task loss, where $\hat{b}_{l}$ are predictions on the $l$-th task. Note that there are a total of $L$ tasks. In our case, $L=2$ as we have speed and torque predictions as the additional tasks.
The third term is the feature mimicking loss, which we use L2 loss. Transformation operations are represented as  $\Phi(\cdot)$ and $\Psi(\cdot)$.
The parameters $\alpha_{l}$ and $\beta_{k}$ balance the influence of multi-task loss and feature mimicking loss on the final prediction task.

%\begin{figure}[htb]
%  \centering
%  \includegraphics[scale=0.7]{./figures/feature_map.pdf}
%  \caption{Visualization of the feature maps of PSPNet our network will learn. (High-level features of PSPNet)}
%  \label{fig6}
%\end{figure}

%\vspace{0.10cm}
%\noindent\textbf{Discussion}: \cavan{discuss the advantage of heterogeneous feature mimicking here.}

%\subsection{Mimicking path and target feature size}

%Table~\ref{tab:feature_dimension} records the mimicking path and target feature size. Note that for PSPNet, we empirically found that selecting channels that are semantically relevant to the task is more effective than performing average pooling over all channels. In our implementation, from the logits of PSPNet we choose two channels that correspond to road and vehicle classes, and perform element-wise maximum operation on all channels to get the third channel. With appropriate resizing, we obtain $30 \times 30 \times 3$ output dimensions for the high-level mimicking path of PSPNet. As to FlowNet, we just perform average pooling over all channels and resize these features to get the target features.

%-----------------------------------
\subsection{Network Initialization and Multi-Stage Training}
\label{subsec:initialization_training}
%-----------------------------------

From our experiments, we found that a good initialization is important to the convergence of a very deep 3D CNN since the parameters of the network are significantly more in comparison to a 2D network.
In our experiments, without a proper initialization, the FM-Net (50-layer 3D ResNet) suffers from convergence problem and even yields a result poorer than a shallow network.
In this work, we follow~\cite{carreira2017quo} to initialize our network. Specifically, we first load the weights of a 2D ResNet-50 model that has been pre-trained on ImageNet~\cite{krizhevsky2012imagenet} in our 3D network.
We then copy the weights of a $w \times w$ kernel $w$ times along the time dimension and normalize the weights by $w$ so that a sequence of video frames will get the same response as it goes through a 2D network. Besides, the stride of both convolution layers and max-pooling layers is set as 1 in the temporal dimension so that the input sequence length does not decrease.

After initialization, we train the network based on Eqn.~\eqref{eqn:final_loss}.
We found that a multi-stage training scheme works well in practice. In particular, the loss defined in Eqn.~\eqref{eqn:final_loss} contains three terms. In the first stage, we optimize \algorithmname~based on the first two terms $\mathcal{L}_\mathrm{steer}$ and $\mathcal{L}_\mathrm{multi}$. In the second stage, we train the network by using all the terms including $\mathcal{L}_\mathrm{mimic}$.
Introducing $\mathcal{L}_\mathrm{mimic}$ at the very beginning of the training yields slightly inferior results to the proposed two-stage strategy.
We conjecture that feature mimicking from heterogeneous networks (\ie, optimizing against $\mathcal{L}_\mathrm{mimic}$) a relatively harder and more complex task in comparison to learning steering angles, speed, and torque. Thus the main network should behave relatively well in steering angle prediction before performing feature mimicking, else feature mimicking would be less efficient.

%////////////////////////////////
\section{Experiments}
\label{sec:experiments}
%////////////////////////////////
% !TEX root = ../deepdrive.tex

\noindent\textbf{Datasets}. We perform evaluations on two standard benchmarks widely-used in the community, namely Udacity~\cite{udacity} and Comma.ai~\cite{santana2016learning} for evaluation.
They are the largest steering angle prediction datasets by far. Note that the Berkeley Deep Drive (BDD) dataset~\cite{yu2018bdd100k} provides vehicle turning directions (i.e., go straight, stop, turn left / right) instead of steering wheel angles. Nonetheless, we conducted experiments on this dataset and provide the results. We show that heterogeneous feature mimicking still helps even in a much larger dataset (BDD dataset provides more than 7 M video frames).

The Udacity dataset is mainly composed of video frames taken from urban roads. It provides a total number of 404,916 video frames for training and 5,614 video frames for testing. This dataset is challenging due to severe lighting changes, sharp road curves and busy traffic.
The images of Comma.ai dataset are mainly captured from highway and urban roads. It contains 11 video clips within 7.5 hours driving. Busy traffic conditions make this dataset challenging.
Note that there is no official and publicly available partition setting for this dataset. For fair comparisons, we benchmark our method and variants using a common setting. Specifically, we use 5\% of each of the 11 clips for validation and testing, chosen randomly as a continuous chunk. The remaining frames are used for training. As pre-processing, we downsample the clips by a factor of two in time and remove video frames whose speed is less than 15 m/s to discard erroneous steering readings. As a result, we obtain a total number of 341,663 frames for training and 23,642 frames for testing. We release the data partitions on our project page\footnote{Project page: \emph{https://cardwing.github.io/projects/FM-Net}}.
Some typical frames of these two datasets are shown in Fig.~\ref{fig:qualitative_results}.

%\begin{figure}[t]
%  \centering
%  \includegraphics[width=\linewidth]{./figures/dataset_images.pdf}
%  \vskip -0.3cm
%  \caption{Example frames from Udacity~\cite{udacity} and Commai.ai~\cite{santana2016learning}. The ground-truth steering angle is indicated by a red line.}\label{dataset_images}
%\end{figure}

\vspace{0.1cm}
\noindent\textbf{Implementation details}. To facilitate the training of our network, the pixel values of input video frames are normalized to lie in [-1, 1]. Frames in Udacity are resized to 160~$\times$~160.
We follow the common practice~\cite{udacity} to use video clips of 10 frames each as inputs (\ie, $N=10$).
The balancing parameters $\alpha_l$ is set as 1.0 for both the speed and torque prediction tasks, while $\beta_k$ is set as 0.2 for all auxiliary networks.
%
%In order to make full use of the data, we randomly select a sequence of 10 consecutive video frames from the total training set.
In Udacity, three vehicle states, \ie, steering angle, torque and speed are used as targets, while we only use steering angle and speed in Comma.ai, since it does not provide steering torque.
A training batch for our network contains 16 video clips. The learning rate is set as $10^{-4}$ in first 30 training episodes and reduced to $10^{-6}$ thereafter.

\vspace{0.1cm}
\noindent\textbf{Evaluation metrics}. We follow existing studies and use mean absolute error (MAE) and root mean square error (RMSE) as metrics.

%-----------------------------------
\subsection{Comparative Evaluations}
\label{exp:comparison}
%-----------------------------------

We compare the proposed method with state-of-the-art approaches on two publicly available datsets, \ie, Udacity and Comma.ai. The results are summarized in Table~\ref{tab:comparison}.

\noindent
\textbf{Udacity}. We compare with~\cite{kim2017interpretable} based on the results reported in their paper. We obtain the results of~\cite{udacity} by using its codes shared by the top team on official Udacity GitHub\footnote{https://github.com/udacity/self-driving-car/.}. All baselines use the same train/test partition. The baseline 3D CNN + LSTM~\cite{udacity} is the best existing method on this dataset.
It is evident from Table~\ref{tab:comparison} that a deeper model (our 50-layer ResNet) is advantageous than 3D CNN. In particular, the MAE is reduced from 2.5598 to 1.9167, a relative improvement of 25\%.
Adding LSTM to model the temporal information further improves the result of 3D ResNet from MAE of 1.9167 to 1.7147.
The best result is yielded by the proposed FM-Net, which is based on 3D ResNet + LSTM but further enhanced with heterogeneous feature mimicking.

\noindent
\textbf{Comma.ai}. Making comparisons on this dataset is more challenging as there are no official or publicly available train/test partition settings. Owing to this reason, we did not include the results reported by Kim \etal~\cite{kim2017interpretable} in Table~\ref{tab:comparison} to avoid unfair comparison. Based on our own partition setting, we run the code of the best baseline in Udacity, \ie, 3D CNN + LSTM~\cite{udacity}, on this data and report its results. Again, 3D ResNet and 3D ResNet + LSTM outperform the shallower baselines, and FM-Net with heterogeneous feature mimicking achieves the best performance. Noticeably, with feature mimicking, we bring down the MAE by 12\% (from 0.7989 to 0.7048), which is significant.

\noindent
\textbf{BDD100K}. As can be seen from Table~\ref{tab:comparison}, it is apparent that feature mimicking can bring considerable performance gains to 3D ResNet + LSTM and outperforms all previous algorithms. The results validate the effectiveness of our proposed feature mimicking method.

\begin{figure*}
  \centering
  \includegraphics[width=0.9\linewidth]{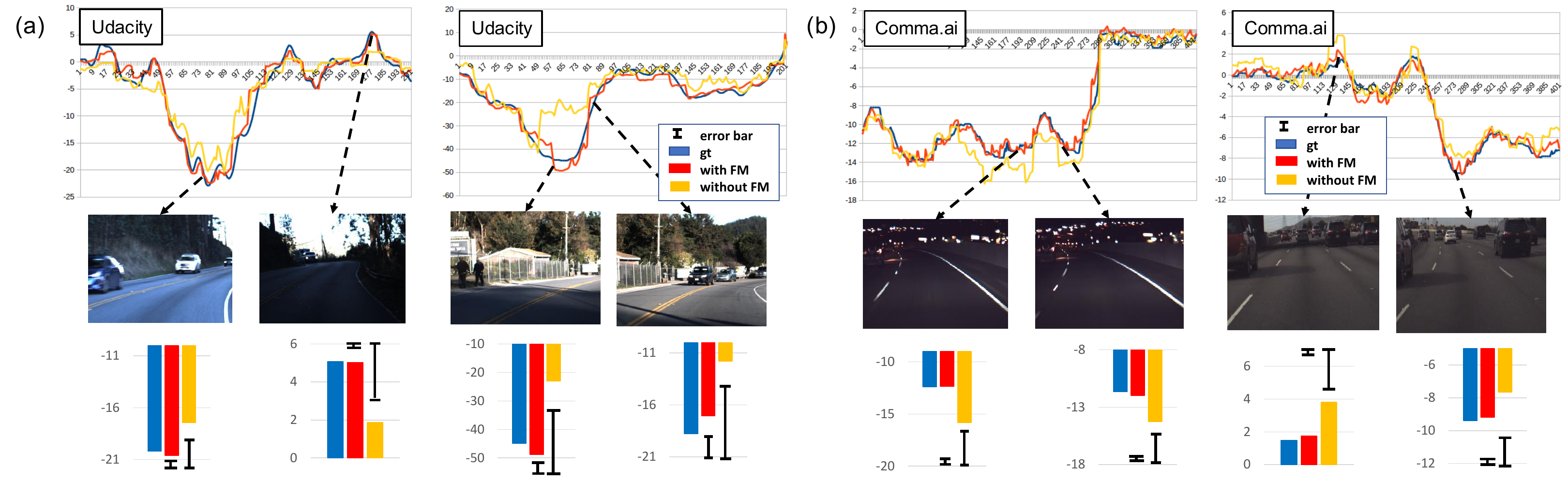}
  \vskip -0.3cm
  \caption{Qualitative results of FM-Net with and without heterogeneous feature mimicking (FM) on (a) Udacity and (b) Comma.ai test sets. The top row shows steering angles over time (x axis denotes frame and y axis represents steering angle). Ground-truth (gt) is represented in blue line. We selected a few representative examples to highlight the advantages of using feature mimicking. The bar charts in the second row provide closer observations on the predictions and errors (represented by the error bar).}
  \label{fig:qualitative_results}
\end{figure*}

\noindent
\textbf{Qualitative results}.
We show qualitative results of FM-Net with and without feature mimicking in Fig.~\ref{fig:qualitative_results}.
It can be observed that FM-Net with heterogeneous feature mimicking yields much stable manoeuvre in driving albeit challenging road conditions including curved roads and extremely dark scenes. The observations are consistent on both Udacity and Comma.ai datasets. %More results are provided in the \textit{supplementary material}.

\begin{table}[t]
\centering
\caption{Comparison with state-of-the-art methods on Udacity, Comma.ai and BDD100K datasets. $\dag$ indicates the results are copied from~\cite{kim2017interpretable}. Note that the proposed FM-Net is based on 3D ResNet+LSTM but further enhanced with heterogeneous feature mimicking.}
\label{tab:comparison}
\footnotesize{
\begin{tabular}{l|c|c}
\hline
\multirow{2}*{Method} & \multicolumn{2}{|c}{Udacity}  \\
\cline{2-3}
~ & MAE & RMSE  \\
\hline \hline
CNN + FCN$\dag$~\cite{bojarski2016end}& 4.1200 & 4.8300  \\
\hline
CNN + LSTM~\cite{kim2017interpretable}& 4.1500 & 4.9300 \\
\hline
CNN + Attention~\cite{kim2017interpretable} & 4.1500 & 4.9300 \\
\hline
3D CNN~\cite{udacity} & 2.5598 & 3.6646 \\
\hline
3D CNN+LSTM~\cite{udacity} & 1.8612 & 2.7167  \\
\hline \hline
3D ResNet~(ours) & 1.9167 & 2.8532 \\
\hline
3D ResNet+LSTM~(ours) & 1.7147 & 2.4899  \\
\hline
\textbf{\algorithmname~(ours)} & \textbf{1.6236} & \textbf{2.3549} \\
\hline
\end{tabular}\vspace{3ex}
\begin{tabular}{l|c|c}
\hline
\multirow{2}*{Method} & \multicolumn{2}{|c}{Comma.ai}  \\
\cline{2-3}
~ & MAE & RMSE  \\
\hline \hline
3D CNN~\cite{udacity} & 1.7539 & 2.7316 \\
\hline
3D CNN + LSTM~\cite{udacity} & 1.4716 & 1.8397  \\
\hline \hline
3D ResNet~(ours) & 1.5427 & 2.4288 \\
\hline
3D ResNet+LSTM~(ours) & 0.7989 & 1.1519  \\
\hline
\textbf{\algorithmname~(ours)} & \textbf{0.7048} & \textbf{0.9831} \\
\hline
\end{tabular}\vspace{3ex}
\begin{tabular}{l|c}
\hline
Method & Accuracy on BDD100K \\
\hline \hline 
FCN + LSTM~\cite{xu2017end} & 82.03\% \\
\hline
CNN + LSTM~\cite{xu2017end} & 81.23\% \\
\hline
3D CNN + LSTM~\cite{udacity} & 82.94\% \\
\hline \hline
3D ResNet + LSTM (ours) & 83.69\% \\
\hline
\textbf{\algorithmname~(ours)} & \textbf{85.03\%} \\
\hline
\end{tabular}
}
\end{table}

%-----------------------------------
\subsection{Ablation Study}
\label{exp:ablation_study}
%-----------------------------------

\begin{table}
\centering
\caption{Performance comparison of mimicking features from different layers of auxiliary networks.  We use ``P'' and ``F'' to denote PSPNet and FlowNet, respectively. The abbreviation is used along with ``L'' and ``M'' and ``H'' to represent \tuple{low}{middle}{high}. For instance, PSPNet feature mimicking (high-level) is abbreviated as ``PH''. Full feature mimicking means ``PH + PM + PL + FH + FM + FL''.}
\label{tab:fm_ablation}
\footnotesize{
\begin{tabular}{l|c|c|c|c}
\hline
\multirow{2}*{Method} & \multicolumn{2}{|c|}{Udacity} & \multicolumn{2}{|c}{Comma.ai} \\
\cline{2-5}
~ & MAE & RMSE & MAE & RMSE \\
\hline \hline
Without feat. mimick & 1.7147 & 2.4899 & 0.7989 & 1.1519 \\
\hline
PH + FH & 1.6826 & 2.4013 & 0.7514 & 1.0836 \\
\hline
PM + FM & 1.6928 & 2.4659 & 0.7749 & 1.0836 \\
\hline
PL + FL & 1.6869 & 2.4521 & 0.7627 & 1.1204 \\
\hline
PH + PM + PL & 1.6653 & 2.3847 & 0.7315 & 1.0574 \\
\hline
FH + FM + FL & 1.6573 & 2.3746 & 0.7259 & 1.0215 \\
\hline
With full feat. mimick & \textbf{1.6236} & \textbf{2.3549} & \textbf{0.7048} & \textbf{0.9831} \\
\hline
\end{tabular}
}
\end{table}

\vspace{0.15cm}
\noindent\textbf{The effectiveness of heterogeneous feature mimicking}.
We summarize the performance of mimicking features from different layers of auxiliary networks in Table.~\ref{tab:fm_ablation}. We have a few observations.
(1) Feature mimicking is beneficial since the one without feature mimicking results in the lowest performance.
(2) High-level feature mimicking (PH+FH) brings slightly more benefits than mid- (PM+FM) and low-level (PL+FL) feature mimicking judging from RMSE. This may be explained from the observation that high-level features contain more semantical meanings than those of mid- and low-levels, as supported by the feature embeddings shown in Fig.~\ref{fig:overview}(b). Another reason could be that it is more fruitful to regularize the high-level features of FM-Net rather than the mid- and low-level features.
(3) Comparing PH + PM + PL and FH + FM + FL, we see no obvious difference between using either PSPNet or FlowNet2 as the sole auxiliary network although the results obtained from using FlowNet2 is marginally better.
(4) The best performance is achieved when we use both auxiliary networks and activate the \tuple{low}{middle}{high} mimicking paths.
Mimicking features at different levels from different networks help FM-Net to capture more diverse contextual information, \eg~object motion and scene structure, at different feature resolutions.

\begin{figure}[t]
  \centering
  \includegraphics[width=\linewidth]{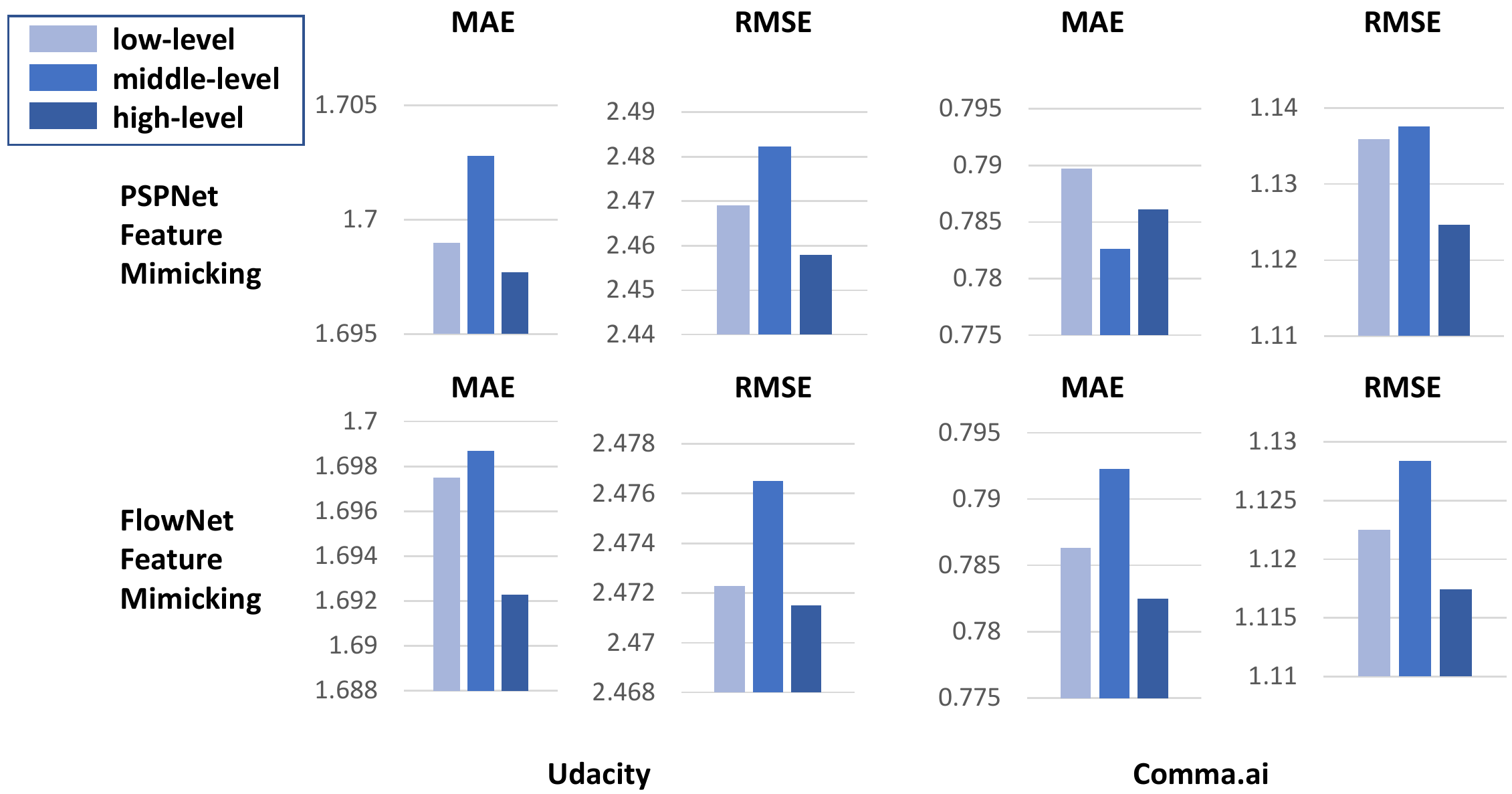}
  \vskip -0.3cm
  \caption{Comparative results of activating a single mimicking path chosen from \tuple{low}{middle}{high} of an auxiliary network.}
\label{fig:different_level_mimicking}
\end{figure}

Table~\ref{tab:fm_ablation} studies some representative combinations of mimicking paths. Next, we further examine the performance of FM-Net when we only allow a single mimicking path chosen from \tuple{low}{middle}{high} of an auxiliary network. The results are shown in Fig.~\ref{fig:different_level_mimicking}.
It is observed that high-level mimicking paths are generally superior to the mid- and low-level paths. It is interesting to learn that regularizing low-level features with mimicking brings more benefits than mid-level features do. This is an intriguing observation that worths further investigations.

\vspace{0.15cm}
\noindent\textbf{Feature mimicking v.s. pre-training}.
Pre-training is an alternative approach to introduce a side task indirectly without performing annotations on the target set -- we can pre-train the FM-Net using the same image segmentation task on Cityscape dataset~\cite{cordts2016cityscapes} as in the PSPNet and subsequently fine-tune FM-Net on the steering angle prediction task. In this way, we wish to observe if the network could still benefit from the segmentation task.
We compare feature mimicking with this approach and report the results in Table~\ref{tab:compare_with_pretraining}. We include FM-Net without both Cityscape pre-training\footnote{The mIOUs of FM-Net with Cityscape pre-training only (ResNet-50, with Large FOV, without data augmentation, ASPP and CRF) in the validation and testing set of Cityscape are 67.2 and 66.4, respectively, which are comparable with the state-of-the-art method~\cite{chen2018deeplab} (ResNet-101, with data augmentation, Large FOV, ASPP and CRF, validation: 71.4, testing: 70.4).} and feature mimicking as a baseline. In this comparison, the FM-Net variant with feature mimicking only mimics features from PSPNet (\ie, PH+PM+PL in Table~\ref{tab:fm_ablation}). All three methods in Table~\ref{tab:fm_ablation} used ImageNet initialization.
As can be observed, pre-training only yields very marginal improvement. By contrast, feature mimicking brings a higher gain to FM-Net. The results suggest that this na\"{i}ve pre-training scheme may not be the most effective way in our problem context: (1) the side task pre-training employs Cityscape, which introduces a domain gap when we applied the pre-trained network on Udacity and Comma.ai; (2) the network structure of FM-Net is not optimal for direct learning from the image segmentation task.
Feature mimicking alleviates the two aforementioned issues as it approximates PSPNet's features by using target data as input. And it focuses to approximate \tuple{low}{middle}{high} features well rather do well on the segmentation tasks itself, thus network architecture becomes a less crucial issue.

\begin{table}[t]
\centering
\caption{Comparing heterogeneous feature mimicking and network pre-training. Variants (A) without both Cityscape pre-training and feature mimicking, (B) with Cityscape pre-training only, and (C) with feature mimicking only.}
\label{tab:compare_with_pretraining}
\footnotesize{
\begin{tabular}{l|c|c|c|c}
\hline
\multirow{2}*{FM-Net} & \multicolumn{2}{|c|}{Udacity} & \multicolumn{2}{|c}{Comma.ai} \\
\cline{2-5}
~ & MAE & RMSE & MAE & RMSE \\
\hline \hline
Variant A & 1.7147 & 2.4899 & 0.7989 & 1.1519 \\
\hline
Variant B & 1.7125 & 2.4614 & 0.7842 & 1.0908 \\
\hline
Variant C & 1.6653 & 2.3847 & 0.7315 & 1.0574 \\
\hline
\end{tabular}
}
\end{table}

%////////////////////////////////
\section{Conclusion}
%////////////////////////////////

Contextual learning from side networks is a meaningful exploration not attempted before. We have presented a novel scheme of training very deep 3D CNN for the task on end-to-end steering angle prediction. Specifically, we found that approximating multi-level features from heterogeneous auxiliary networks provide strong supervisory signals and regularization to the main network. In our experiments, we have shown that PSPNet and FlowNet2 help our FM-Net to learn better in capturing contextual information such as scene structure and object motion. With the proposed heterogeneous feature mimicking, the proposed FM-Net achieves a new state-of-the-art on both Udacity and Comma.ai benchmarks.

\noindent\textbf{Acknowledgement}: This work is supported by SenseTime Group Limited and the General Research Fund sponsored by the Research Grants Council of the Hong Kong SAR (CUHK 14241716, 14224316. 14209217).

\bibliographystyle{aaai}
\bibliography{DeepDrive}
\end{document}